\documentclass[review,11pt,authoryear]{elsarticle}
\usepackage{natbib}
\usepackage{amssymb}
\usepackage[dvips]{graphicx}
\usepackage{amsmath}
\usepackage{amsfonts}
\usepackage{amsthm}
\usepackage{pifont}
\usepackage{txfonts}
\usepackage{xcolor}
\usepackage{pst-plot}
\usepackage{endfloat}

\usepackage[colorlinks=true]{hyperref}
\newcommand{\longvec}[1]{\stackrel\longrightarrow{\smash{#1}\vphantom{i}\,}}

\newcommand{\marginparred}[1]{\marginpar{\color{red}{#1}}} 
\journal{Ergonomics}

\begin{document}

\begin{frontmatter}

\title{{\color{red}Determination of subject-specific muscle fatigue rates under static fatiguing operations}}

\author[TSINGHUA]{Liang MA\corref{cor}}
\ead{liangma@tsinghua.edu.cn; liang.ma.thu@gmail.com}
\author[TSINGHUA]{Wei ZHANG}
\author[TSINGHUA]{Bo HU}
\author[IRCCyN]{Damien CHABLAT}
\author[IRCCyN,ECN]{Fouad BENNIS}
\author[EADS]{Fran\c{c}ois GUILLAUME}
\address[TSINGHUA]{Department of Industrial Engineering, Tsinghua University, 100084, Beijing, P.R.China}
\address[IRCCyN]{ Institut de Recherche en Communications et en Cybern\'{e}tique de Nantes, UMR 6597  \\IRCCyN - 1, rue de la No\"{e}, BP 92 101 - 44321 Nantes CEDEX 03, France}
\address[ECN]{ \'{E}cole Centrale de Nantes, 1, rue de la No\"{e}, BP 92 101 - 44321 Nantes CEDEX 03, France}
\address[EADS]{EADS Innovation Works, 12, rue Pasteur - BP 76, 92152 Suresnes Cedex - France}

\cortext[cor]{Corresponding author: Tel:+86-10-62792665; Fax:+86-10-62794399}

\begin{abstract}

Cumulative local muscle fatigue may lead to potential musculoskeletal disorder (MSD) risks {\color{red}, and subject-specific muscle fatigability needs to be considered to reduce potential MSD risks.} This study was conducted to determine local muscle fatigue rate at shoulder joint level based on an exponential function derived from a muscle fatigue model. Forty male subjects participated in a fatiguing operation under a static posture with a range of relative force levels (14\% - 33\%). Remaining maximum muscle strengths were measured after different fatiguing sessions. The time course of strength decline was fitted to the exponential function. Subject-specific fatigue rates of shoulder joint moment strength were determined. Good correspondence ($R^2>0.8$) was found in the regression of the majority (35 out of 40 subjects). Substantial inter-individual variability in fatigue rate was found and discussed.

\end{abstract}


\end{frontmatter}

\textbf{Practitioner summary}

{\color{red}Different workers have different muscle fatigue attributes. Determination of joint-level subject-specific muscle fatigue rates can facilitate physical task assignment, work/rest scheduling, MSD prevention, and worker training and selection}. \marginparred{R2Q1}

\section{Introduction}

Human intervention is often involved in occupational activities, especially in material handling, assembly, and maintenance tasks \citep{melhorn2001successful, melhorn2001management, kumar2001theories}. In those activities, the operator needs sufficient muscle strength to meet force requirement for operating equipment or sustaining external loads. {\color{red}Insufficient strength can lead to overexertion of the musculoskeletal system and to consequent injuries \citep{ARMSTRONG1993,Chaffin1999}}.\marginparred{R1Q1} 

A decrease in muscle strength is often experienced in a physical operation under a sub-maximal force, either in a continuous way or in an intermittent way \citep{wood1997mfd, Yassierli2009effects}. This decrease in maximum force output results from different sources, such as muscle fatigue, musculoskeletal disorders, lack of motivation, etc. Among those sources, muscle fatigue is one of the most prevalent reasons and is defined as ``any exercise-induced reduction in the capacity to generate force or power output'' \citep{vollestad1997mhm}. {\color{red}Muscle fatigue exposes operator to more risks of overexertion, and cumulative muscle fatigue may result in musculoskeletal disorders (MSDs) \citep{ARMSTRONG1993,Chaffin1999}.}\marginparred{R1Q2} 

Fatigue progression is closely dependent on task assignment and subject-specific fatigue attributes. Different task parameters (load, duration of exertion, posture and motion, etc.) lead to different fatigue progressions in physical operations \citep{chaffin2009evolving,  Yassierli2009effects, enoka2012muscle}. Individual physical attributes (e.g., strength, fatigue rates, recovery rates, etc.) can influence the fatigue progression as well. It is believed that fatigue attributes differentiate from each other among operators \citep{yassierli2007influence, Yassierli2009effects, avin2010sex, avin2011age}.  Determination of subject-specific fatigue attributes is of interest and of importance for physical work design \citep{chaffin2009evolving}. 

Muscle fatigue progression has been studied mainly from two different approaches. One approach is maximum endurance time (MET) approach. MET can assess the ability to resist fatigue by measuring the maximal duration while exerting a force at a specific level until failure. A large amount of effort has been contributed to developing MET models for different muscle groups under different static working conditions \citep{rohmert1960ees, rohmert1973pdr, rohmert1986ssn, bishu1995fer, kanemura1999evf, mathiassen1999psf, Garg2002, law2010endurance}. Although the MET models can predict the maximum endurance time under a given relative force level, the decrease in the muscular strength cannot be predicted directly by MET models. Moreover, due to the nature of the formation in those MET models from group data, it is difficult to determine subject-specific fatigue attributes. 

Another approach to characterise muscle fatigue progression is to develop muscle fatigue models, and hence to predict the strength decline directly. Some work \citep{giat1993musculotendon, ding2000predictive} contributed to complex physiological muscle fatigue models. Those models are able to describe the muscle fatigue progression precisely for a single muscle. However, they are too complex for industrial application due to the difficulty of identifying a great number of parameters in the model. 

Some researchers established some fatigue models by conducting fatiguing tasks \citep{deeb1992exponential, sogaard2006effect, romanliu2004qau, romanliu2005dfc, iridiastadi2006emfa}. Among those models, {\color{red}exponential declines in muscular strength are found\citep{deeb1992exponential, romanliu2004qau, yassierli2007influence}}\marginparred{R2Q9}. However, the fitting parameters in those exponential functions could not implicate more information about fatigue attributes of each subject. 

Some other researchers \citep{liu2002dmm, ma2008nsd, xia2008tam} have tried other models to describe muscle fatigue progression. \marginparred{R1Q3}{\color{red} \citet{xia2008tam} proposed a three-compartment muscle fatigue model based on muscle motor units model in \citet{liu2002dmm}, and they have run simulation to demonstrate fatigue progression under a variety of loading conditions. In this model, fatigue and recovery attributes of different types of muscle fibers were assigned with different values in the simulation. However, the lack of validation limits the application of this model. }

\citet{ma2008nsd}  proposed a muscle fatigue model to describe muscle fatigue progression from a macro perspective. In this model, task parameters and muscle fatigue rate are combined together to understand the fatigue caused by tasks and fatigue attributes. \citet{ma2009gfr} developed an approach to determine fatigue resistances of different muscle groups using this fatigue model. Twenty-four MET models \citep{elahrache2006pvd} for different muscle groups can be effectively {\color{red} fitted and explained} by this approach. {\color{red}We suggest}\marginparred{R2Q10} that this muscle fatigue model is capable of assessing fatigue progression of a muscle group in static cases. Moreover, we found that the muscle fatigue progression of each subject under static cases can be predicted in the form of an exponential function derived from the fatigue model, and one important factor (fatigue rate) in this model emerges to represent subject-specific muscle fatigue attribute.

{\color{red}Regarding the subject-specific fatigability\marginparred{R1Q0}, some other measures were used to assess muscle fatigability, such as endurance time, EMG power spectrum (Median frequency (MF) and Median Power Frequency (MPF)), Mean Arterial Pressure (MAP) and so on \citep{clark2003gd, hunter2004fatigability, hunter2005muscle, yoon2007mechanisms, frey2010endurance, cote2012critical}.  However, as pointed out by \citet{vollestad1997mhm}, the greatest limitation is that those measures are indirect to measure muscle fatigue. Therefore, we chose to use the fatigue rate in \citet{ma2009gfr} to represent inter-individual difference in fatigue progression to beyond those limitations. }

We conducted this {\color{red}study}\marginparred{R1Q4} to verify whether the fatigue progression under a static operation can be well {\color{red}fitted} \marginparred{R2Q11}by a specific exponential function derived from the muscle fatigue model and to check whether the fitting parameter (fatigue rate) could represent subject-specific fatigue attributes among different subjects. This paper is organised as follows: Section \ref{sec:theo} describes the theoretical approach to determine subject-specific fatigue rate based on a muscle fatigue model. Section \ref{sec:matandmet} presents materials, methods, and experiment procedure in this study. Section \ref{sec:results} and Section \ref{sec:discuss} {\color{red}show results and provide discussion}\marginparred{R2Q12, R1Q17}.

\section{Subject-specific fatigue rate determination}
\label{sec:theo}

The purpose of this study was to determine subject-specific muscle group fatigue attributes by analysing muscle fatigue progression based on a theoretical muscle fatigue model. 

{\color{red}\citet{ma2008nsd} proposed a muscle fatigue model in the form of a differential equation (Eq. \ref{eq:FcemDiff})}\marginparred{R1Q6, R2Q13}.  The muscle fatigue model {\color{red}describes}\marginparred{R2Q14} the change of the maximum remaining strength over time. Related parameters and their descriptions are given in Table \ref{tab:Parameters}. In this model, the fatigue rate ($k$) is a parameter to indicate the relative speed of strength decline within a muscle.

\begin{equation}
\label{eq:FcemDiff}
			\frac{dF_{rem}(t)}{dt} = -k \frac{F_{rem}(t)}{MVC}F_{load}(t)
\end{equation}

\begin{table*}[htp]
	\centering
	\caption{Parameters in the dynamic fatigue model}
	\label{tab:Parameters}
		\begin{tabular}{lcp{0.7\textwidth}}
		\hline
		Item & Unit & Description\\
		\hline
		$MVC$	or $F_{max}$	& N 				&	Maximum voluntary muscle strength under nonfatigued state\\
		$F_{rem}(t)$ 				& N 				& Maximum voluntary remaining muscle strength at time $t$\\
		$F_{load}(t)$				& N 				& External load that the muscle needs to bear\\
		$k$									& min$^{-1}$ 	& Fatigue rate\\
		$\%MVC$							&						&Percentage of the voluntary maximum contraction\\
		$f_{MVC}$						&						&$\%MVC/100$, $f_{MVC}=\frac{F_{load}}{MVC}$.\\
		\hline			
		\end{tabular}
\end{table*}

{\color{red}In a static muscular operation}\marginparred{R1Q7}, $F_{load}(t)$ keeps constant, and the reduction of the muscular strength can be further predicted by Eq. \ref{eq:static}. This equation describes the muscle fatigue progression in {\color{red} the form of an exponential function}\marginparred{R2Q13}. Three parameters ($F_{max}$, $F_{load}$, and $k$) act upon the fatigue progression under a static operation. In general, $F_{load}$ is determined by the task design, and it can be measured or calculated via force analysis, and $F_{max}$ and $F_{rem}$ can be measured to unfold the muscle fatigue progression. The fatigue rate $k$ is task independent and it is influenced by several factors (e.g., muscle fiber type composition, age, and gender) \citep{ma2009gfr}.\marginparred{R1Q12} 

{\color{red}
\begin{equation}
	\label{eq:static}
	\dfrac{F_{rem}(t)}{F_{max}}=e^{-k\,f_{MVC}\;t}
\end{equation}}\marginparred{R1Q8}

According to the definition of muscle fatigue, muscle fatigue progression can also be described by measuring the maximum remaining muscle strengths over time during a fatiguing operation. If the same muscle progression can be depicted using both ways, it will suggest {\color{red} the parameter $k$ could be determined using the muscle fatigue model. } {\color{red}Therefore, it is essential to verify whether the fatigue progression of each subject under a static fatiguing operation follows an exponential function in the form of Eq. \ref{eq:static} with a high coefficient of determination $R^2$.}\marginparred{R1Q9, R1Q10, R1Q11,R2Q6}

Suppose that we have already a set of real measurements for a given static operation, where $F_{t_{i}}$ indicates the maximum remaining strength $F_{rem}$ at time instant $t_{i}$. At the beginning of a physical task, the subject is supposed having no muscle fatigue. Therefore, the $F_{t=0}$ can be treated as the maximum voluntary contraction $F_{max}$. Equation \ref{eq:static} can be further transformed into Eq. \ref{eq:curvefit}. 

\begin{equation}
\label{eq:curvefit}
	\dfrac{\ln\left(\dfrac{F_{t_{i}}}{MVC}\right)}{f_{MVC}}=-k\;t_i
\end{equation}

Since $f_{load}$ and $MVC$ are measured and/or known, fatigue rate $k$ can be further determined by linear regression. A high goodness of fit between maximum remaining strengths and the exponential function would suggest the usefulness of the fatigue rate.  The goodness of fit is assessed by the $R^2$ value in a linear regression without an intercept.

\section{Materials and Methods}
\label{sec:matandmet}

\subsection{Subjects}
Since the focus of this study is on manufacturing and assembly and the majority of the operators are male workers, 40 right-handed male industrial workers participated in the experiment after signing a written informed consent. The age, stature, body mass, upper limb anthropometry data, and body mass index (BMI) were recorded or measured upon arrival at the laboratory (see Table \ref{tab:participants}). Participation was limited to individuals with no previous history of upper limb problems. Ethical approval for this study was obtained from the human research ethical advisory committee of Tsinghua University.
 
\begin{table*}[htp]
     \centering
     \caption{Subject physical characteristics}
         \begin{tabular}{lcccc}
             \hline
             Characteristic&Mean& Standard Deviation (SD) & Maximum&Minimum\\
             \hline
             Age (year)&41.2&11.4&58&19\\
             Height (cm)&171.2&5.1&183.0&160.0\\
             Mass (kg)&70.2&10.4&95.0&50.0\\
             Upper limb (cm)&23.6&3.0&31.0&16.0\\
             Lower limb (cm)&25.6&1.8&29.0&22.0\\
             BMI & 23.9 & 3.35 & 31.1 & 18.7  \\
             \hline
         \end{tabular}
     \label{tab:participants}
\end{table*}

\subsection{Task design}
\begin{figure}[htbp]
	\centering
		\includegraphics[width=0.60\textwidth]{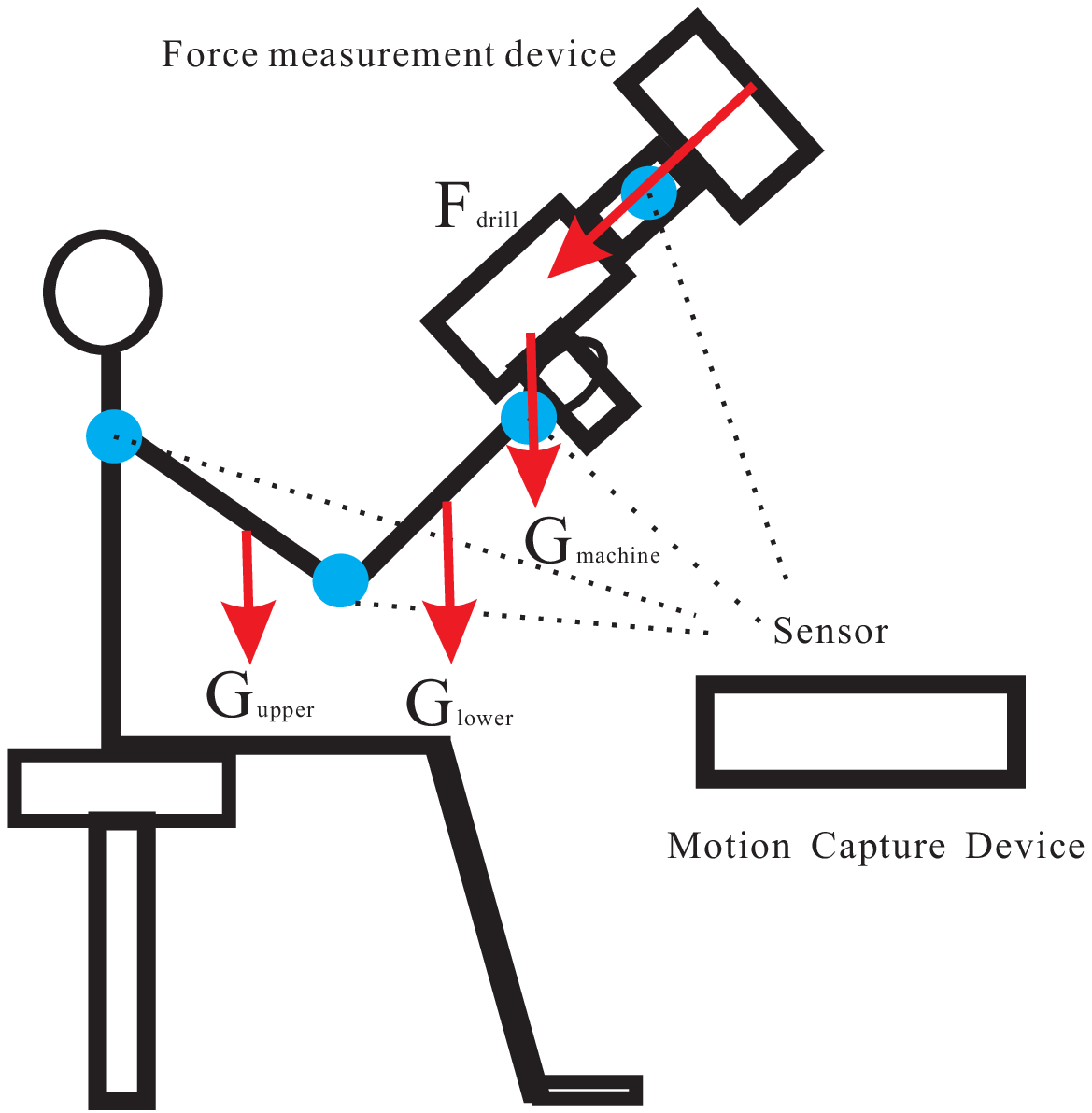}
	\caption{Seated static posture in the experiment and materials used in the experiment. $F_{drill}$: drilling force; $G_{machine}$: the weight of the drilling machine; $G_{upper}$: the weight of the upper arm; $G_{lower}$: the weight of the lower arm}
	\label{fig:drilling}
\end{figure}

In this study, we used a typical overhead drilling operation under laboratory conditions to measure muscle fatigue progression at shoulder joint level (see Fig. \ref{fig:drilling}). This task was simplified from a real drilling operation from the program of the European Aeronautic Defence and Space (EADS). This operation was selected as a typical task because there are a few ergonomics {\color{red}problems} \citep{melhorn2001successful}\marginparred{R2Q15}. A heavy external load demands great physical strength to hold a machine and maintain the operation for a certain period, and local muscle fatigue occurs rapidly in upper limb and lower back. MSD risks can be increased by force overexertion and long-lasting vibration while drilling. 

The magnitude of the external load and the duration of the operation are two key factors to simulate the real situation. The relative load needs to be adequate so that subjects can experience fatigue and can endure the operation for a certain period. In this case, according to the strength model in \citet{Chaffin1999} and the MET models in \citet{elahrache2006pvd}, a subject must apply a drilling force of 25 N and hold a drilling machine with a mass of 2.5 kg. The drilling force is only applied along the drilling direction towards the subject. The estimated moment generated by the external load (including the weight of the arm) is about 33\% of the shoulder joint flexion strength of a 50$^{th}$ percentile male and the endurance time under this load is estimated around 4 minutes.

\subsection{Measures}

\begin{figure}[htp]
\centering
\begin{pspicture}(0,0)(9,5)
\def\beam{
        \pspolygon[linewidth=1.5pt](0,0)(3,0)(3,1)(0,1)
        \pspolygon[linewidth=1.5pt](3,0.2)(3.2,0.2)(3.2,0.8)(3,0.8)
      	\psline[linewidth=1pt,linestyle=dashed]{-}(0,0.5)(3.5,0.5)
      	\rput[l]{-14}(3.3,0.2){Force measurement}
      	\rput[tl]{-14}(0,0){\;Beam}
        }
\rput{14}(5.125,3.52){\beam}
\psline[linewidth=1pt,linestyle=dashed]{-}(5,4)(6.2,4)
\psarc(5,4){1}{0}{14}
\rput[bl](6.3,4){$14.5^{\circ}$}
\psline[linewidth=1pt]{*-*}(1,4)(2,1)
\psline[linewidth=1pt]{*-*}(2,1)(4,2.5)
\psline[linewidth=1pt]{*-*}(4,2.5)(5,4)
\psline[linewidth=1pt]{*->}(1.5,2.5)(1.5,1)
\psline[linewidth=1pt]{*->}(3,1.75)(3,0.25)
\psline[linewidth=1pt]{*->}(4.5,3.25)(4.5,1.75)
\psline[linewidth=1pt]{*->}(5,4)(3.5,3.5)
\psline[linewidth=1pt,linestyle=dashed]{-}(1,4)(1,2.8)
\psarc(1,4){1}{270}{288.4} 
\psline[linewidth=1pt,linestyle=dashed]{-}(2,1)(2.33,0)
\psarc(2,1){0.5}{288.4}{36.8} 
\rput[tl](1,2.8){$q_1$ }
\rput[tl](2.40,0.7){$q_2$ }
\rput[br](1,4){\em S}
\rput[tr](2,1){\em E} 
\rput[br](4,2.5){\em W}
\rput[br](5,4){\em  D} \rput[br](1.5,1){$\mathbf{G}_u$ }
\rput[bl](3,0.25){ $\mathbf{G}_f$ } \rput[br](4.5,1.75){$\mathbf{G}_m$ }
\rput[br](3.5,3.5){$\mathbf{F}_{d}$}
\end{pspicture}
\caption{Force schema in the drilling operation. S: shoulder; E: elbow; W: wrist; D: drilling point. $\mathbf{F}_d$: drilling force applied to the upper limb; $\mathbf{G}_m$: weight of the drilling machine; $\mathbf{G}_u$: weight of the upper arm; $\mathbf{G}_f$: weight of the forearm; $q_1$: flexion of the upper arm; and $q_2$: flexion of the elbow} \label{fig:momentarm}
\end{figure}
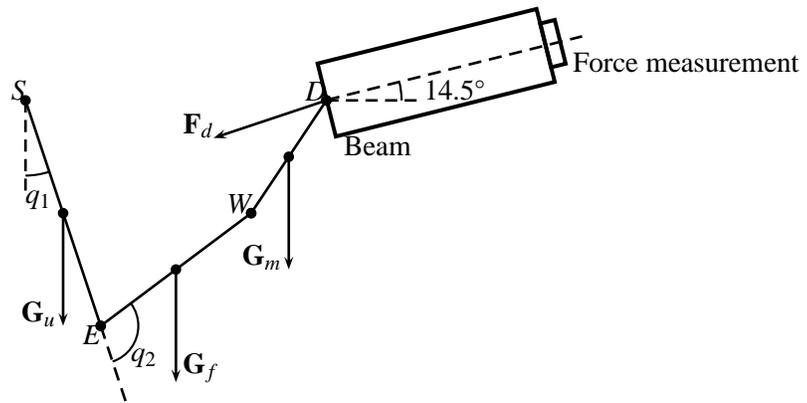

{\color{red}In this study, we used shoulder joint moment strength to describe fatigue progression and measured the maximum force output in the drilling direction to estimate shoulder joint moment strength (see Fig. \ref{fig:momentarm}). }\marginparred{R1Q5}

We assumed that the measured force was determined by joint moment strength of the right upper limb. Shoulder joint and elbow joint have similar strength profiles according to the joint moment strength models \citep{Chaffin1999}, and shoulder joint has higher fatigability in MET models than elbow joint \citep{frey2010endurance}. Furthermore, it was obvious that shoulder joint was charged with a much larger moment load than elbow joint in this drilling case. Therefore, the bottleneck for the output strength was shoulder joint. 

The moment load in shoulder joint can be approximately estimated by Eq. \ref{eq:moment} (see Fig. \ref{fig:momentarm}).The mass and the centre of gravity of each body segment were estimated from the anthropometry database \citep{Chaffin1999}. 

\begin{equation}
	\label{eq:moment}
	\begin{aligned}
	\mathbf{\Gamma}_{load}&=\left(\dfrac{\mathbf{s}-\mathbf{e}}{2}\right)\times \mathbf{G}_u +\left(\dfrac{\mathbf{w}+\mathbf{e}}{2}-\mathbf{s}\right)\times \mathbf{G}_f\\
	&	+ \left(\dfrac{\mathbf{d}+\mathbf{w}}{2}-\mathbf{s}\right)\times \mathbf{G}_m
	+\left(\mathbf{d}-\mathbf{s}\right)\times \mathbf{F}_d
	\end{aligned}
\end{equation}
where $\mathbf{s}$, $\mathbf{e}$, $\mathbf{w}$, and $\mathbf{d}$ represent the coordinates of the positioning sensors attached to the shoulder (S), elbow (E), wrist (W), and drilling contact point (D), respectively. {\color{red}In our experiment, since the subject's arm is strictly limited within the sagittal plane, we just measured the force and calculated the torque within this plane.\marginparred{R1Q16}}

Since we use joint moment strength, the fatigue model in Eq. \ref{eq:FcemDiff} can be changed to Eq. \ref{eq:MomentDiff} by replacing all of the force terms with joint moment terms. 
\begin{equation}
\label{eq:MomentDiff}
			\frac{d\Gamma_{rem}(t)}{dt} = -k \frac{\Gamma_{load}(t)}{\Gamma_{max}}\Gamma_{rem}(t)
\end{equation}

Under this simplified case, $\Gamma_{load}$ can be determined by force analysis, and $f_{MVC}$ can be calculated from data analysis by normalizing the $\Gamma_{load}$ over $\Gamma_{max}$. $\Gamma_{t_i}$ is a joint maximum remaining strength, and it can be estimated by measuring the maximum remaining force $F_{t_i}$ in the drilling direction. 

\subsection{Material}
In the experiment, we want to measure the muscular strength in the drilling direction and to estimate the joint moment strength around the shoulder joint. Therefore, force measurement and motion capture devices were used (see Figure \ref{fig:drilling}). 

{\color{red}We used a dynamometer to measure the drilling force in the drilling direction (see Figure \ref{fig:drilling}). The dynamometer measures the pressing force perpendicular to the load cell surface with a measurement range upto 300 $N$ and a precision of 1 $N$. }\marginparred{R1Q13}

We use the magnetic motion capture device FASTRAK\textsuperscript{\textregistered} (POLHEMUS Inc.) to capture the upper limb posture in the experiment. As shown in Figure \ref{fig:drilling}, we attached four positioning sensors to the key joints of upper limb and the drilling machine. We captured the Cartesian coordinates of the shoulder, the elbow, the wrist, and the contact point between the drilling machine and the work piece. The tracking device runs at 30 $Hz$ per sensor and has a static position accuracy of 1 $mm$. We used the recorded coordinates of each tracker to reconstruct the posture of the worker in post-experiment analysis. 

We provided the drilling force with a wooden beam with a mass of 10 $kg$. We used wooden material to avoid magnetic distortions caused by ferrous material and to ensure motion capture accuracy. We suspended the wooden beam with an inclination angle between the beam and the horizontal line of $14.5^{\circ}$. During operation, the subject had to push the beam against the force measurement device and hold it for a certain period. According to the force analysis of the pendulum, a tangential force of 25 $N$ was charged to the upper limb. Before each trial, we calibrated this external load to ensure that there was exactly a force of 25 $N$ applied in the drilling direction. We provided the weight of the drilling machine with a drilling tool made from concrete with a mass of 2.5 $kg$.

\subsection{Experiment Procedure}
\label{sec:experiment}

Each subject had to complete ten sessions: one MVC session and nine fatiguing sessions. 

In the MVC session, maximum voluntary contraction ($MVC$) was determined as the greatest exerted force in the drilling direction over three trials. In each trial, we verbally encouraged subjects to maintain the maximum force peak for three to five seconds. The measured $MVC$ was also denoted as $F_0$ to represent the subject's initial maximum strength at the beginning of the operation. Between each trial, subjects were asked to take at least a 5-minute rest until self-reported full recovery \citep{Chaffin1999}.

There were nine fatiguing sessions with different time intervals (15, 30, 45, 60, 75, 90, 120, 150, and 180 seconds). The sequence to complete those nine sessions was randomly assigned for each subject. In each fatiguing session, subject was asked to hold the constant external load for the time interval $t_i$ (e.g., 30 sec). After that, the remaining muscle strength $F_{t_{i}}$ (e.g., $F_{30}$) was measured by asking the subject to exert the maximal voluntary strength with a force peak from three to five seconds.  After the measurement, subjects took a rest for at least five minutes or even longer until self-reported total recovery. 

After the recovery, the subject was asked randomly to {\color{red} conduct a MVC trial\marginparred{R2Q2}. Full recovery would be recognized if the measured maximum strength in this trial was more than 95\% of the MVC. Otherwise, the subject would be asked to take longer break until full recovery}\marginparred{R1Q15}. Once subject reported that he could not sustain the operation within the session, experiment would be stopped immediately to avoid injuries to subject. 

Within each session, subject was seated upright, and right shoulder was fixed to a shoulder support against the wall to restrict the movement of shoulder and to decrease the engagement of lower back. Left upper limb was free, and right upper limb was limited in the sagittal plane by position constraints. The position constraints provided posture references to subject to maintain the initial posture as well as possible, but provided no support to upper limb. The seated height and location was adjusted according to subject’s height and upper limb length to reduce variability among the different subjects.\marginparred{R1Q14}

\subsection{Data Analysis}
The objective of this analysis was twofold. \marginparred{R2Q16}
{\color{red}
\begin{enumerate}
	\item to test if the fatigue progression of each subject in shoulder joint maximum strength can be well fitted by the exponential function derived from the muscle fatigue model or not\marginparred{R2Q3};
	\item to analyse the relationship between muscle fatigue rate and joint moment strength.
\end{enumerate}
} 

For the first objective, we fitted muscle fatigue progression of each subject with the exponential function (Eq. \ref{eq:curvefit}). The coefficient of determination of each fitting was recorded for each subject and analysed. For the second objective, we selected two groups of subjects to assess the relationship of joint moment strength and joint fatigue rate, {\color{red}since muscles engaged in the action mainly determine maximum joint strength and we assumed that determinant muscle-related factors for muscle strength could probably act effects on muscle fatigue rate as well.}  One group (Group A) is the subjects with 10 highest joint moment strengths; another group is the subjects with 10 lowest joint moment strengths. {\color{red}Besides joint strength, some other measures (e.g., BMI, age) may also influence fatigue rate. Correlations between fatigue rate and other measures were also statistically analysed. We used SPSS to do all the statistical analysis and fitting.}\marginparred{R1Q12, R1Q17, R1Q19}

\section{Results}
\label{sec:results}

\subsection{Fatigue progression}

\begin{figure}[htbp]
	\centering
		\includegraphics[width=0.80\textwidth]{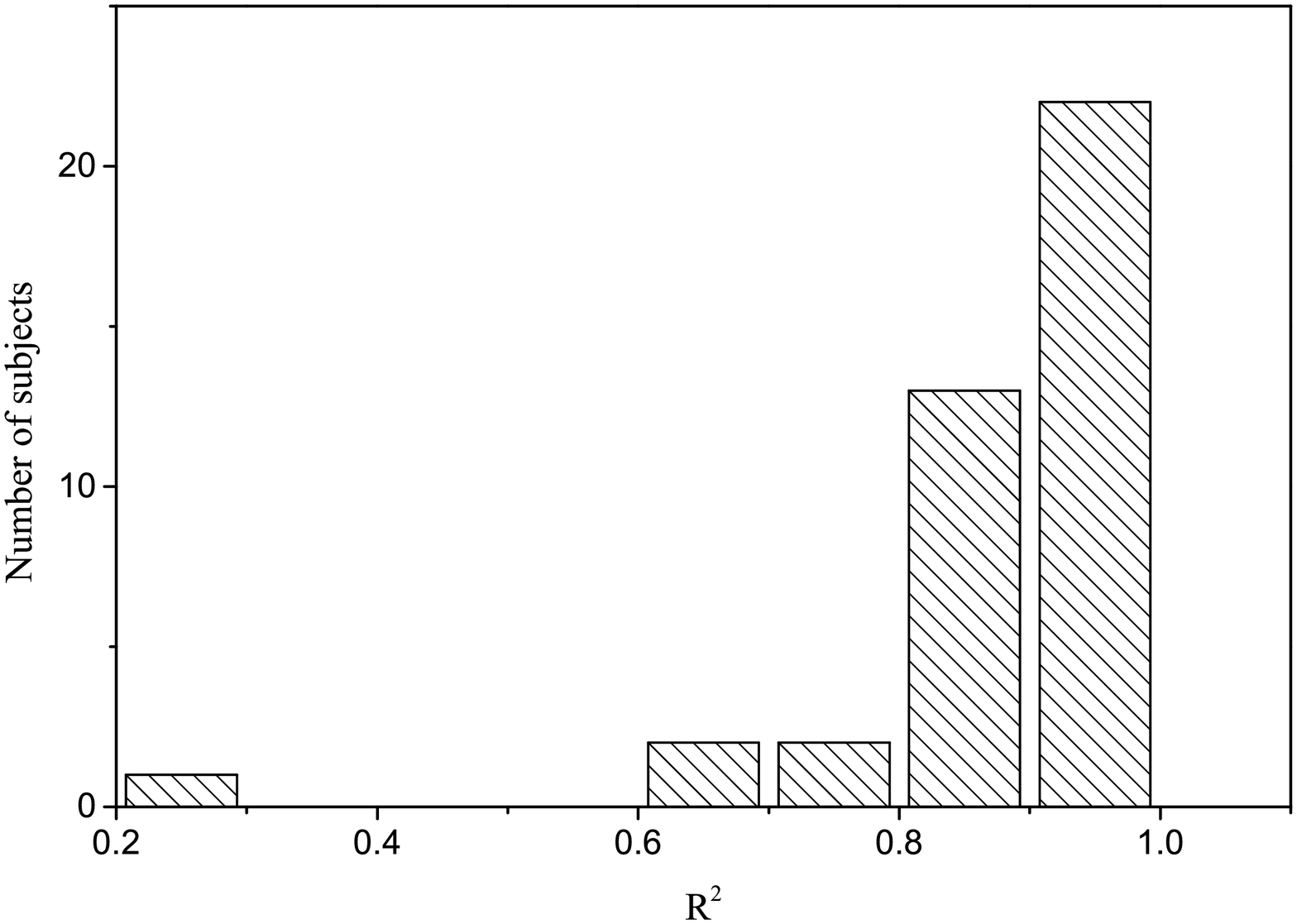}
	\caption{Histogram of coefficient of determination $R^2$}
	\label{fig:rsquare}
\end{figure}

\begin{figure}[htbp]
	\centering
		\includegraphics[width=0.80\textwidth]{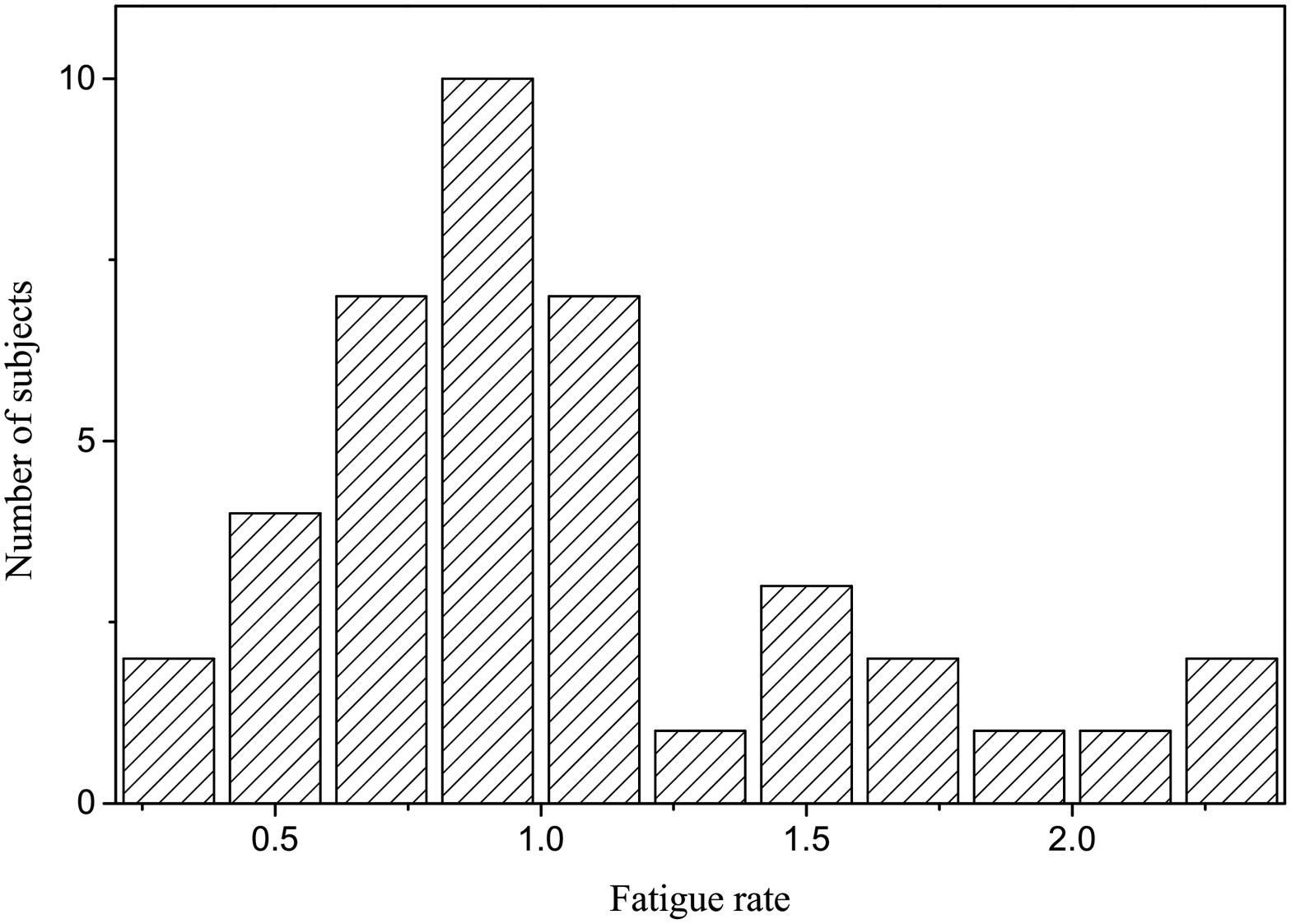}
	\caption{Histogram of fatigue rate at shoulder joint}
	\label{fig:fatiguerate}
\end{figure}

Joint moments of each subject were normalized over the estimated maximum moment strength. Then the normalized values were fitted using Eq. \ref{eq:curvefit} to calculate the fitting coefficient $R^2$ and to determine the fatigue rate. The statistical results of $R^2$ of the regression and the fatigue rate $k$ were listed in Table \ref{tab:fatigueratio}, and the histograms of $R^2$ and $k$ were shown in Fig. \ref{fig:rsquare} and Fig. \ref{fig:fatiguerate}.

{\color{red}35 out of 40 subjects had a high coefficient of determination $R^2$ over 0.8 in joint moment regression. Four out of the other five subjects had a fair coefficient $R^2$ over 0.63, and only one of the five subjects had a very poor $R^2$ (0.23). }\marginparred{R2Q4}

\begin{table*}[htp]
	\centering
		\caption{Statistical analysis of fatigue rate $k$}
	\label{tab:fatigueratio}
		\begin{tabular}{rcccc}
		\hline
		Item & Mean & SD & Minimum & Maximum\\
		\hline
		Joint moment strength ($Nm$)& 45.1 & 7.4 & 67.4 & 32.1\\
		$k$	&	1.02	&0.49	&	0.37	&	2.29	\\
		$R^2$&0.87&0.14 &	0.23	&	0.99\\
		\hline			
		\end{tabular}
\end{table*}\marginparred{R2Q18}

\marginparred{R1Q18}

\marginparred{R1Q18}

\subsection{Relationship between fatigue rate and joint maximum strength}

A pair-wise correlation matrix was determined between joint moment fatigue rate, joint moment strength, age and BMI. The results were shown in Table \ref{tab:corrmatrix}. It was found that joint moment strength is strongly correlated with fatigue rate ($p<0.05$), while no strong correlations were found in the pair of fatigue rate and BMI and the pair of fatigue rate and age.

\begin{table}[htbp]
  \centering
 \caption{Correlation matrix for study variables.( $^*p<0.05$)}
  \begin{tabular}{lllll}
    \hline
                                       &Fatigue rate&    BMI &Joint moment strength & Age\\
\hline
                    Fatigue rate&1                 & -0.09 &0.616$^*$                            &0.033\\
                                 BMI&                   &1         &0.09                               &0.40$^*$  \\
    Joint moment strength&                  &            &1                                  &0.072\\
                                 Age&                   &           &                                    &1\\
    \hline
  \end{tabular}
 
  \label{tab:corrmatrix}
\end{table} \marginparred{R1Q19}

No significant differences were found between two groups in age (Group A: 42.9 (SD=7.4); Group B: 39.1 (SD=15.4), $p=0.49$) and BMI (Group A: 24.9 (SD=2.4); Group B: 25.2 (SD=4.2) $p=0.78$). Significant differences ($p=0.00$) were found in joint maximum strength between Group A (mean=60.8 $Nm$ (SD=4.7$Nm$)) and Group B (Mean=37.7 $Nm$ (SD=3.3 $Nm$)) .\marginparred{R1Q20}

With use of t-test, Table \ref{tab:correlation} showed the difference of fatigue rates between different groups. The subjects with higher strength have significantly higher fatigue rate {\color{red}even though} \marginparred{R2Q21}the relative load is smaller than the subjects with lower strength. 
 
\begin{table}[htbp]
	\centering
	\caption{Effect of muscle strength on fatigue rate}
			\begin{tabular}{lllll}
		\hline
		Group  & Mean &  SD & $t$& $p$-Value\\
		\hline
		$k$\\
		\hline
		A  & 1.47 &0.53&4.628&0.0001\\
		B & 0.64 &0.20\\
	\hline
		\end{tabular}

	\label{tab:correlation}
\end{table}\marginparred{R2Q18}

\subsection{Posture change during the drilling operation}
\label{sec:posture}
\begin{figure}[htp]
	\centering
		\includegraphics[width=0.9\textwidth]{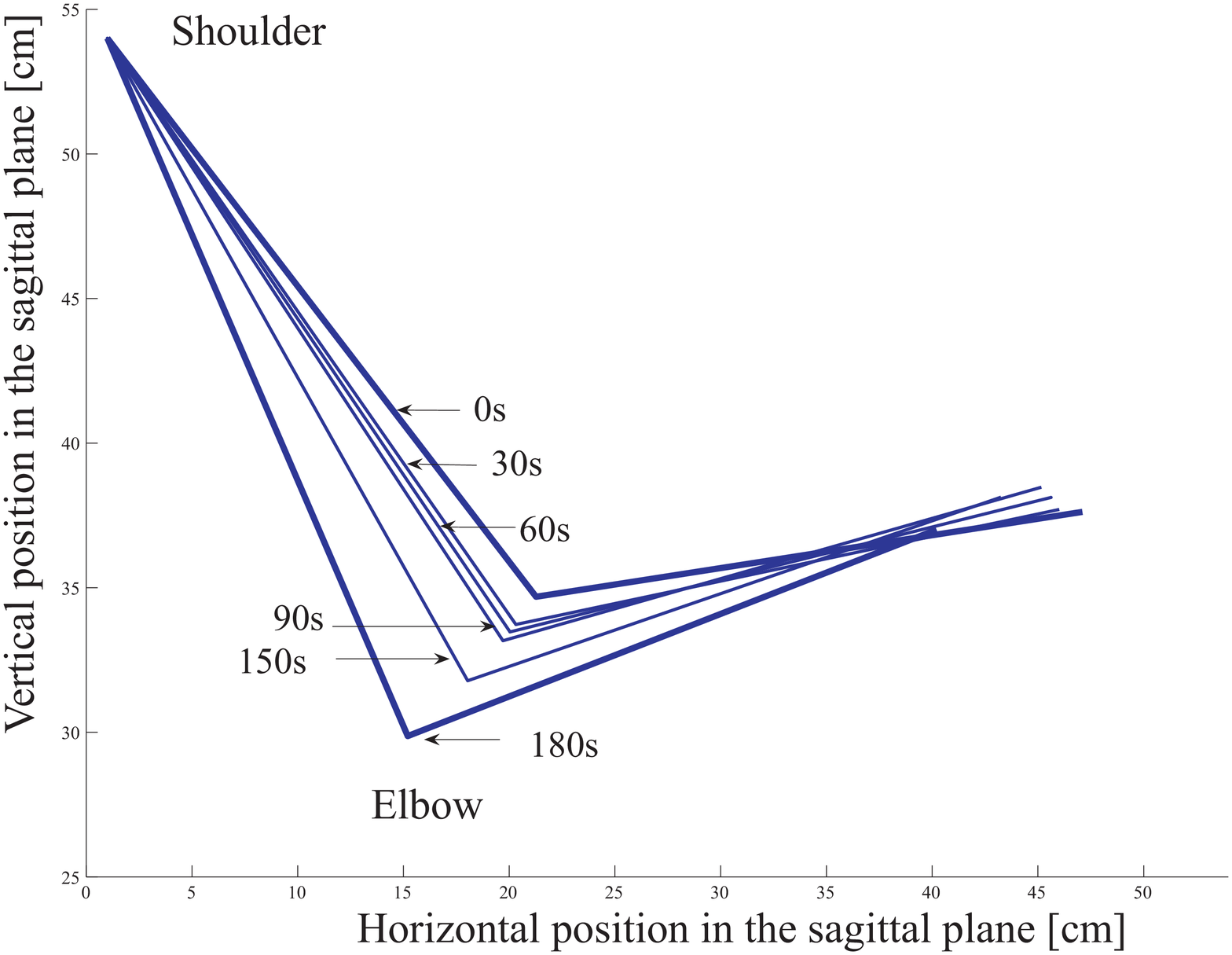}
	\caption{Posture change during the drilling operation (mean values across participants)}
	\label{fig:posture}
\end{figure}

We calculated the posture of upper limb during the drilling operation from the motion data. Because the arm was constrained in the sagittal plane, only the flexion angles of shoulder joint and elbow joint were calculated to represent arm posture to eliminate influence from different limb lengths. The statistical results of elbow flexion and shoulder flexion angles across participants were calculated and shown in Table \ref{tab:posture}. The posture change during the working process is shown in Figure \ref{fig:posture}. The changes in the posture followed the following tendency: the greater the fatigue was, the closer the upper limb was to the trunk. In this way, the moment produced by the mass of the upper limb around shoulder joint could be reduced.

\begin{table*}[htp]
	\centering
	\small
	\caption{Posture change during the experiment (deg)}
	\label{tab:posture}
			\begin{tabular}{lcccccccccc}
		\hline
		Time (sec)&0&15&30&45&60&75&90&120&150&180\\
		\hline
		Elbow\\
		\hline
		Mean &50.1&53.1&55.1&55.1&57.5&59.9&59.9&64.2&66.7&75.5\\
		SD   &16.1&15.4&15.0&15.7&16.4&19.0&19.2&19.9&21.3&21.9\\
		\hline
		Shoulder\\
		\hline
		Mean&46.4&44.5&43.6&44.2&42.8&42.1&41.9&39.7&37.5&30.5\\
		SD   &16.2&15.0&14.6&15.2&14.7&16.6&17.0&16.6&17.9&17.3\\	
		\hline		
		\end{tabular}
\end{table*}

\section{Discussion}
\label{sec:discuss}

\subsection{Muscle fatigue progression}
In this study, we found that the fatigue progression at shoulder joint among most of the subjects (35/40) can be well fitted ($R^2>0.8$) by the muscle fatigue model. Five out of 40 subjects were found with low $R^2$ coefficients. \marginparred{R2Q6} 

There are some reasons leading to those relatively poor fittings.  First, the motivation and the attitude of the subject during the experiment could influence the result. Second, {\color{red} even though}\marginparred{R2Q17} the posture of the arm was strictly constrained in the sagittal plane, subjects could still have a certain degree of mobility. The willingness to maintain the posture may influence the muscle recruitment strategy, muscle coordination, and the posture during the experiment. {\color{red} Third, weak muscular strength could probably lead to poor fitting performance indirectly.\marginparred{R2Q20} The subject with a $R^2=0.23$ had the second lowest MVC among the 40 subjects. Lower muscle strength means relatively higher physical workload during the experiment and higher demand to maintain the posture, and hence the static operation would be more possibly aversely influenced due to the unwillingness to maintain the posture. }\marginparred{R2Q7, R2Q19}

Fatigue may occur at any step along the pathway that is involved in muscle contraction \citep{berne2004physiology}. Both the metabolic factors within the muscle and the impairment of activation could contribute to the decline in muscle power output \citep{Chaffin1999,allen2008skeletal}. In a physical operation, muscle fatigue progression could be influenced by physical task, motivation, and individual fatigue attributes\marginparred{R1Q21}. However, under static operation and effective verbal encouragement, the influences from motivation could be rather limited. Therefore, muscle fatigue progression was probably mainly caused by relative loads and subject-specific fatigue attributes in this study. 

The fatigue model  (Eq. \ref{eq:static}) is formed from a macro perspective and can be explained based on motor units principle from a micro perspective \citep{ma2008nsd}.{\color{red} The product of relative load and fatigue rate determines the decline of muscle strength.}  \marginparred{R1Q23} According to muscle physiology, fatigue rate can be recognized as a parameter to represent the overall fatigue resistant performance of a muscle group at joint level under a specific task \citep{ma2009gfr}.

We selected this model to determine fatigue attribute for the following reasons: (1) in comparison to Deeb's model\citep{deeb1992exponential}, this model enables us to decouple relative load and subject-specific attribute; (2) {\color{red}muscle fatigue rate can be influenced by muscle composition, neuromuscular activation patterns, and coordination among single muscles, therefore the fatigue rate in this model could cover more effects of influencing factors than the fatigue rates of different types of muscle fibers in \citet{xia2008tam}}; \marginparred{R1Q22}(3)this model is less complex than  the three-compartment model in \citet{xia2008tam} , and it could be practical for industry application. 

\subsection{Subject-specific fatigue rate}

We found that there are substantial differences among different subjects in fatigue rates at shoulder joint level. It suggests that the fatigue rate $k$ can be used to characterise different fatigue attributes among subjects at shoulder joint level under this specific fatiguing condition.

The underlying mechanisms for those differences in fatigue rate are very complex. Physiologically, differences in fatigue rates are mainly caused by four factors: (1) muscle strength (muscle mass) and associated vascular occlusion, (2) substrate utilisation, (3) muscle composition, and (4) neuromuscular activation patterns \citep{hicks2001sdh, ma2009gfr}. At the same time, demographic parameters (age, gender) and their interactions can lead to changes in muscle composition as well \citep{mademli2008eva, Yassierli2009effects}. In addition, personal working experience or physical exercises and living style (e.g., smoking) can also change the muscle strength and endurance via the adaptive response of muscle cells to regular external loads \citep{berne2004physiology, wust2008skeletal}. All those factors generate effects together on subject-specific fatigue attributes. \marginparred{R1Q23}

We found also that {\color{red}fatigue rates are positively correlated to maximum joint moment strength in this study, even though the relative loads for the subjects with higher strength were lower.} \marginparred{R2Q17, R2Q21} Between-subject differences in the ratio of type I muscle fibers (slow twitch, more fatigue resistant) to type II muscle fibers (fast twitch, less fatigue resistant) might explain the differences in fatigue rates. Muscle strength depends strongly on muscle fiber size and muscle fiber composition \citep{fitts1991determinants}. Subjects in Group A and Group B did not have significant differences in BMI and age, which implicates that the strength differences were probably not mainly caused by muscle mass or muscle fiber size or age-related factors. It could be concluded that the strength differences were caused by different compositions of muscle fiber types. Subjects with higher strengths could probably have a higher proportion of type II muscle fibers and lower proportion of type I muscle fibers. That leads to higher joint moment strength and faster fatigability in the muscle.

We did not find significant correlation between fatigue rate and age and between fatigue rate and BMI. Regarding the age effect, we did not control our subjects strictly to two age groups. The subjects in this experiment were in their young age or middle age, which might not enough to reveal the aging effect. Regarding the BMI, most of the subjects were in normal weight and overweight group. Only a few subjects belong to Class I obesity or underweight category.

\subsection{Posture change}
Although posture references were provided to avoid mismatches in different test trials, it was still very difficult for subjects to maintain the purely static posture during the operation. Small changes occurred in the experiment, but those changes did not generate excessive variation in joint strength. In our case, the variation of the maximum joint moment strength is no more than 3\% relative to the maximum strength under the initial posture according to the joint moment strength model \citep{Chaffin1999}. The change of the joint strength due to posture change might lead change of the relative strength. We checked the sensitivity of the change, and the changed maximum joint strength would lead to no more than{\color{red} 4\% change of the relative strength ($4\%f_{MVC}$), which was acceptable in this case.} \marginparred{R1Q20}

Several reasons may cause posture change in the experiment. Fatigue might be one of those reasons. Changes in the posture can be explained by a global posture control strategy, which includes decreasing the joint loads in the operation by moving the upper limb closer to the body; a similar finding has been reported by \citet{fuller2008pmc}. However, that change would influence joint strength\citep{romanliu2005uls, anderson2007mvj}. {\color{red}Besides fatigue, there were still other error sources leading to the change of the posture. First, the actual posture was determined by the anthropometry of different subjects. Different arm lengths could cause potential differences in elbow flexion and shoulder flexion. Second, the posture was calculated from the position sensors attached to the key joints. Each subject might have different sensor configurations, which might lead to calculation errors. Third, there might be differences among the postures that each subject took in different fatiguing sessions. }\marginparred{R2Q5, R1Q24, R1Q25, R1Q20}

\subsection{Limitations}
\marginparred{R1Q26, R2Q22, R2Q23}
{\color{red}
There are several limitations in this study. First, the focus of the present study is on the fatigue effect in static industrial operations in a continuous working process. Recovery effect is not taken into consideration in this study. Second, this study was conducted under a simplified overhead drilling operation, and the conclusion drawn from this study has a strong task dependency. A simplified overhead drilling operation decreases the reliability of applying the findings into industry. Some other MSD causes, such as vibration \citep{kattel1999effects} were neglected from this study. Third, the force analysis in this study is available only for a static case. In a real operation, the motion involved in the operation could result in a different dynamic workload. Moreover, only fatigue with the relative force falling from 14\% to 33\% (Mean=24.3\%, SD=4.4\%) of the specific job operation was tested, so the result that was obtained is available only for similar physical operations. Last but not least, the fatigue progression was measured under static isometric contraction, and the results could not be extended for dynamic or intermittent fatiguing tasks. }

\section{Conclusions and perspectives}
This paper provides an experimental approach to determine subject-specific fatigue rate at shoulder joint level. Fatigue progression in a simplified static drilling operation was measured and analysed using an exponential muscle fatigue model. The muscle fatigue progression in joint moment strength could be well fitted by the fatigue model ($R^2>0.8$). This result suggests that the muscle fatigue model could be used to describe fatigue progression for industrial operations within the range of 14\% -33\% of the relative submaximal level under static cases. Different fatigue rates among subjects could be used to characterise the individual fatigue attributes under the same workload.  Determination of subject-specific fatigue rates could be useful for physical task assignment, worker training, worker selection and work design. 

{\color{red}
Since fatigue rate is important and it could be influenced by a number of factors, further study would be necessary to find the effects of these influencing factors \citep{cote2012critical}. Gender difference, age difference, and joint difference in fatigue rates could be investigated. Static continuous and dynamic intermittent tasks could be investigated under different relative load levels.  More strict posture control is necessary to eliminate effects of posture change.}\marginparred{R2Q8}

\section*{Acknowledgments}
\addcontentsline{toc}{section}{Acknowledgments}
This research was supported by the EADS and the R\'{e}gion des Pays de la Loire (France) in the context of collaboration between the \'{E}cole Centrale de Nantes (Nantes, France) and Tsinghua University (Beijing, PR China). This research was supported by the National Natural Science Foundation of China as well(Grant number: NSFC71101079).

\addcontentsline{toc}{section}{References}
\marginparred{R2Q24}
\bibliographystyle{elsarticle-harv}
\bibliography{ergonomics}

\begin{thebibliography}{49}
\expandafter\ifx\csname natexlab\endcsname\relax\def\natexlab#1{#1}\fi
\expandafter\ifx\csname url\endcsname\relax
  \def\url#1{\texttt{#1}}\fi
\expandafter\ifx\csname urlprefix\endcsname\relax\def\urlprefix{URL }\fi

\bibitem[{Allen et~al.(2008)Allen, Lamb, and Westerblad}]{allen2008skeletal}
Allen, D.~G., Lamb, G., Westerblad, H., 2008. Skeletal muscle fatigue: cellular
  mechanisms. Physiological reviews 88~(1), 287--332.

\bibitem[{Anderson et~al.(2007)Anderson, Madigan, and
  Nussbaum}]{anderson2007mvj}
Anderson, D., Madigan, M., Nussbaum, M., 2007. {Maximum voluntary joint torque
  as a function of joint angle and angular velocity: Model development and
  application to the lower limb}. Journal of Biomechanics 40~(14), 3105--3113.

\bibitem[{Armstrong et~al.(1993)Armstrong, Buckle, Fine, Hagberg, Jonsson,
  Kilborn, Silverstein, Sjogaard, and Viikari-Juntura}]{ARMSTRONG1993}
Armstrong, T., Buckle, P., Fine, L.~J., Hagberg, M., Jonsson, B., Kilborn, A.,
  Silverstein, B.~A., Sjogaard, G., Viikari-Juntura, 1993. A conceptual model
  for work-related neck and upper-limb musculoskeletal disorders. Scandinavian
  Journal of Work, Environment and Health 19~(2), 73--84.

\bibitem[{Avin et~al.(2010)Avin, Naughton, Ford, Moore, Monitto-Webber, Stark,
  Gentile, Frey, and Laura}]{avin2010sex}
Avin, K., Naughton, M., Ford, B., Moore, H., Monitto-Webber, M., Stark, A.,
  Gentile, A., Frey, L., Laura, A., 2010. Sex differences in fatigue resistance
  are muscle group dependent. Medicine \& Science in Sports \& Exercise
  42~(10), 1943--1950.

\bibitem[{Avin and Law(2011)}]{avin2011age}
Avin, K.~G., Law, L. A.~F., 2011. Age-related differences in muscle fatigue
  vary by contraction type: a meta-analysis. Physical Therapy 91~(8),
  1153--1165.

\bibitem[{Berne et~al.(2004)Berne, Levy, Koeppen, and
  Stanton}]{berne2004physiology}
Berne, R., Levy, M.~N., Koeppen, B.~M., Stanton, B., 2004. Physiology. Elsevier
  Inc.

\bibitem[{Bishu et~al.(1995)Bishu, Kim, and Klute}]{bishu1995fer}
Bishu, R., Kim, B., Klute, G., 1995. {Force-endurance relationship: does it
  matter if gloves are donned?} Applied Ergonomics 26~(3), 179--185.

\bibitem[{Chaffin(2009)}]{chaffin2009evolving}
Chaffin, D., 2009. The evolving role of biomechanics in prevention of
  overexertion injuries. Ergonomics 52~(1), 3--14.

\bibitem[{Chaffin et~al.(1999)Chaffin, Andersson, and Martin}]{Chaffin1999}
Chaffin, D.~B., Andersson, G. B.~J., Martin, B.~J., 1999. Occupational
  biomechanics, 3rd Edition. Wiley-Interscience.

\bibitem[{Clark et~al.(2003)Clark, Manini, Doldo, and
  Ploutz-Snyder}]{clark2003gd}
Clark, B.~C., Manini, T.~M., Doldo, N.~A., Ploutz-Snyder, L.~L., 2003. Gender
  differences in skeletal muscle fatigability are related to contraction type
  and emg spectral compression. Journal of Applied Physiology 94~(6),
  2263--2272.

\bibitem[{C{\^o}t{\'e}(2012)}]{cote2012critical}
C{\^o}t{\'e}, J.~N., 2012. A critical review on physical factors and functional
  characteristics that may explain a sex/gender difference in work-related
  neck/shoulder disorders. Ergonomics 55~(2), 173--182.

\bibitem[{Deeb et~al.(1992)Deeb, Drury, and Pendergast}]{deeb1992exponential}
Deeb, J., Drury, G., Pendergast, D., 1992. An exponential model of isometric
  muscular fatigue as a function of age and muscle groups. Ergonomics 35~(7-8),
  899--918.

\bibitem[{Ding et~al.(2000)Ding, Wexler, and
  Binder-Macleod}]{ding2000predictive}
Ding, J., Wexler, A.~S., Binder-Macleod, S.~A., 2000. A predictive model of
  fatigue in human skeletal muscles. Journal of Applied Physiology 89~(4),
  1322--1332.

\bibitem[{{El ahrache} et~al.(2006){El ahrache}, Imbeau, and
  Farbos}]{elahrache2006pvd}
{El ahrache}, K., Imbeau, D., Farbos, B., 2006. Percentile values for
  determining maximum endurance times for static muscular work. International
  Journal of Industrial Ergonomics 36~(2), 99--108.

\bibitem[{Enoka(2012)}]{enoka2012muscle}
Enoka, R.~M., 2012. Muscle fatigue--from motor units to clinical symptoms.
  Journal of biomechanics 45~(3), 427--433.

\bibitem[{Fitts et~al.(1991)Fitts, McDonald, and
  Schluter}]{fitts1991determinants}
Fitts, R.~H., McDonald, K.~S., Schluter, J.~M., 1991. The determinants of
  skeletal muscle force and power: their adaptability with changes in activity
  pattern. Journal of biomechanics 24, 111--122.

\bibitem[{Frey~Law and Avin(2010)}]{frey2010endurance}
Frey~Law, L.~A., Avin, K.~G., 2010. Endurance time is joint-specific: a
  modelling and meta-analysis investigation. Ergonomics 53~(1), 109--129.

\bibitem[{Fuller et~al.(2008)Fuller, Lomond, Fung, and
  C{\^o}t{\'e}}]{fuller2008pmc}
Fuller, J., Lomond, K., Fung, J., C{\^o}t{\'e}, J., 2008. {Posture-movement
  changes following repetitive motion-induced shoulder muscle fatigue.} Journal
  of electromyography and kinesiology: official journal of the International
  Society of Electrophysiological Kinesiology.

\bibitem[{Garg et~al.(2002)Garg, Hegmann, Schwoerer, and Kapellusch}]{Garg2002}
Garg, A., Hegmann, K., Schwoerer, B., Kapellusch, J., 2002. {The effect of
  maximum voluntary contraction on endurance times for the shoulder girdle}.
  International Journal of Industrial Ergonomics 30~(2), 103--113.

\bibitem[{Giat et~al.(1993)Giat, Mizrahi, and Levy}]{giat1993musculotendon}
Giat, Y., Mizrahi, J., Levy, M., 1993. A musculotendon model of the fatigue
  profiles of paralyzed quadriceps muscle under fes. Biomedical Engineering,
  IEEE Transactions on 40~(7), 664--674.

\bibitem[{Hicks et~al.(2001)Hicks, Kent-Braun, and Ditor}]{hicks2001sdh}
Hicks, A., Kent-Braun, J., Ditor, D., 2001. Sex differences in human skeletal
  muscle fatigue. Exercise and Sport Sciences Reviews 29~(3), 109--112.

\bibitem[{Hunter et~al.(2005)Hunter, Critchlow, and Enoka}]{hunter2005muscle}
Hunter, S.~K., Critchlow, A., Enoka, R.~M., 2005. Muscle endurance is greater
  for old men compared with strength-matched young men. Journal of applied
  physiology 99~(3), 890--897.

\bibitem[{Hunter et~al.(2004)Hunter, Critchlow, Shin, and
  Enoka}]{hunter2004fatigability}
Hunter, S.~K., Critchlow, A., Shin, I.-S., Enoka, R.~M., 2004. Fatigability of
  the elbow flexor muscles for a sustained submaximal contraction is similar in
  men and women matched for strength. Journal of Applied Physiology 96~(1),
  195--202.

\bibitem[{Iridiastadi and Nussbaum(2006)}]{iridiastadi2006emfa}
Iridiastadi, H., Nussbaum, M., 2006. {Muscle fatigue and endurance during
  repetitive intermittent static efforts: development of prediction models}.
  Ergonomics 49~(4), 344--360.

\bibitem[{Kanemura et~al.(1999)Kanemura, Kobayashi, Hosoda, Minematu, Sasaki,
  Maejima, Tanaka, Matuo, Takayanagi, Maeda, et~al.}]{kanemura1999evf}
Kanemura, N., Kobayashi, R., Hosoda, M., Minematu, A., Sasaki, H., Maejima, H.,
  Tanaka, S., Matuo, A., Takayanagi, K., Maeda, T., et~al., 1999. {Effect of
  visual feedback on muscle endurance in normal subjects}. Journal of Physical
  Therapy Science 11~(1), 25--29.

\bibitem[{Kattel et~al.(1999)}]{kattel1999effects}
Kattel, B., et~al., 1999. The effects of rivet guns on hand-arm vibration.
  International journal of Industrial ergonomics 23~(5-6), 595--608.

\bibitem[{Kumar(2001)}]{kumar2001theories}
Kumar, S., 2001. Theories of musculoskeletal injury causation. Ergonomics
  44~(1), 17--47.

\bibitem[{Law and Avin(2010)}]{law2010endurance}
Law, L., Avin, K., 2010. Endurance time is joint-specific: A modelling and
  meta-analysis investigation. Ergonomics 53~(1), 109--129.

\bibitem[{Liu et~al.(2002)Liu, Brown, and Yue}]{liu2002dmm}
Liu, J., Brown, R., Yue, G., 2002. {A dynamical model of muscle activation,
  fatigue, and recovery}. Biophysical Journal 82~(5), 2344--2359.

\bibitem[{Ma et~al.(2009)Ma, Chablat, Bennis, and Zhang}]{ma2008nsd}
Ma, L., Chablat, D., Bennis, F., Zhang, W., 2009. {A new simple dynamic muscle
  fatigue model and its validation}. International Journal of Industrial
  Ergonomics 39~(1), 211--220.

\bibitem[{Ma et~al.(2011)Ma, Chablat, Bennis, Zhang, Hu, and
  Guillaume}]{ma2009gfr}
Ma, L., Chablat, D., Bennis, F., Zhang, W., Hu, B., Guillaume, F., 2011. {A
  novel approach for determining fatigue resistances of different muscle groups
  in static cases}. International Journal of Industrial Ergonomics 41~(1),
  10--18.

\bibitem[{Mademli and Arampatzis(2008)}]{mademli2008eva}
Mademli, L., Arampatzis, A., 2008. {Effect of voluntary activation on
  age-related muscle fatigue resistance.} Journal of Biomechanics.

\bibitem[{Mathiassen and Ahsberg(1999)}]{mathiassen1999psf}
Mathiassen, S., Ahsberg, E., 1999. {Prediction of shoulder flexion endurance
  from personal factor}. International Journal of Industrial Ergonomics 24~(3),
  315--329.

\bibitem[{Melhorn et~al.(2001{\natexlab{a}})Melhorn, Wilkinson, and
  O'Malley}]{melhorn2001successful}
Melhorn, J., Wilkinson, L., O'Malley, M., 2001{\natexlab{a}}. Successful
  management of musculoskeletal disorders. Human and Ecological Risk Assessment
  7~(7), 1801--1810.

\bibitem[{Melhorn et~al.(2001{\natexlab{b}})Melhorn, Wilkinson, and
  Riggs}]{melhorn2001management}
Melhorn, J., Wilkinson, L., Riggs, J., 2001{\natexlab{b}}. Management of
  musculoskeletal pain in the workplace. Journal of occupational and
  environmental medicine 43~(2), 83--83.

\bibitem[{Rohmert(1960)}]{rohmert1960ees}
Rohmert, W., 1960. {Ermittlung von Erholungspausen f{\"u}r statische Arbeit des
  Menschen}. European Journal of Applied Physiology 18~(2), 123--164.

\bibitem[{Rohmert(1973)}]{rohmert1973pdr}
Rohmert, W., 1973. {Problems in determining rest allowances Part 1: use of
  modern methods to evaluate stress and strain in static muscular work.}
  Applied Ergonomics 4~(2), 91--95.

\bibitem[{Rohmert et~al.(1986)Rohmert, Wangenheim, Mainzer, Zipp, and
  Lesser}]{rohmert1986ssn}
Rohmert, W., Wangenheim, M., Mainzer, J., Zipp, P., Lesser, W., 1986. {A study
  stressing the need for a static postural force model for work analysis}.
  Ergonomics 29~(10), 1235--1249.

\bibitem[{Roman-Liu and Tokarski(2005)}]{romanliu2005uls}
Roman-Liu, D., Tokarski, T., 2005. {Upper limb strength in relation to upper
  limb posture}. International Journal of Industrial Ergonomics 35~(1), 19--31.

\bibitem[{Roman-Liu et~al.(2005)Roman-Liu, Tokarski, and
  Kowalewski}]{romanliu2005dfc}
Roman-Liu, D., Tokarski, T., Kowalewski, R., 2005. {Decrease of force
  capabilities as an index of upper limb fatigue}. Ergonomics 48~(8), 930--948.

\bibitem[{Roman-Liu et~al.(2004)Roman-Liu, Tokarski, and
  W{\'o}jcik}]{romanliu2004qau}
Roman-Liu, D., Tokarski, T., W{\'o}jcik, K., 2004. {Quantitative assessment of
  upper limb muscle fatigue depending on the conditions of repetitive task
  load}. Journal of Electromyography and Kinesiology 14~(6), 671--682.

\bibitem[{S{\o}gaard et~al.(2006)S{\o}gaard, Gandevia, Todd, Petersen, and
  Taylor}]{sogaard2006effect}
S{\o}gaard, K., Gandevia, S., Todd, G., Petersen, N., Taylor, J., 2006. The
  effect of sustained low-intensity contractions on supraspinal fatigue in
  human elbow flexor muscles. The Journal of physiology 573~(2), 511--523.

\bibitem[{V{\o}llestad(1997)}]{vollestad1997mhm}
V{\o}llestad, N., 1997. {Measurement of human muscle fatigue}. Journal of
  Neuroscience Methods 74~(2), 219--227.

\bibitem[{Wood et~al.(1997)Wood, Fisher, and Andres}]{wood1997mfd}
Wood, D., Fisher, D., Andres, R., 1997. {Minimizing fatigue during repetitive
  jobs: optimal work-rest schedules}. Human Factors: The Journal of the Human
  Factors and Ergonomics Society 39~(1), 83--101.

\bibitem[{W{\"u}st et~al.(2008)W{\"u}st, Morse, De~Haan, Rittweger, Jones, and
  Degens}]{wust2008skeletal}
W{\"u}st, R.~C., Morse, C.~I., De~Haan, A., Rittweger, J., Jones, D.~A.,
  Degens, H., 2008. Skeletal muscle properties and fatigue resistance in
  relation to smoking history. European journal of applied physiology 104~(1),
  103--110.

\bibitem[{Xia and Frey~Law(2008)}]{xia2008tam}
Xia, T., Frey~Law, L., 2008. {A theoretical approach for modeling peripheral
  muscle fatigue and recovery}. Journal of Biomechanics 41~(14), 3046--3052.

\bibitem[{Yassierli and Nussbaum(2009)}]{Yassierli2009effects}
Yassierli, Nussbaum, M., 2009. Effects of age, gender, and task parameters on
  fatigue development during intermittent isokinetic torso extensions.
  International Journal of Industrial Ergonomics 39~(1), 185--191.

\bibitem[{Yassierli et~al.(2007)Yassierli, Nussbaum, Iridiastadi, and
  Wojcik}]{yassierli2007influence}
Yassierli, Nussbaum, M., Iridiastadi, H., Wojcik, L., 2007. The influence of
  age on isometric endurance and fatigue is muscle dependent: a study of
  shoulder abduction and torso extension. Ergonomics 50~(1), 26--45.

\bibitem[{Yoon et~al.(2007)Yoon, Schlinder~Delap, Griffith, and
  Hunter}]{yoon2007mechanisms}
Yoon, T., Schlinder~Delap, B., Griffith, E.~E., Hunter, S.~K., 2007. Mechanisms
  of fatigue differ after low-and high-force fatiguing contractions in men and
  women. Muscle \& nerve 36~(4), 515--524.

\end{thebibliography}

\end{document}